\documentclass[10pt,twocolumn,letterpaper]{article}

\usepackage{cvpr}
\usepackage{times}
\usepackage{epsfig}
\usepackage{graphicx}
\usepackage{amsmath}
\usepackage{amssymb}
\usepackage{epstopdf}

\usepackage[pagebackref=true,breaklinks=true,letterpaper=true,colorlinks,bookmarks=false]{hyperref}

\cvprfinalcopy

\def\cvprPaperID{21} 

\begin{document}

\title{Detecting and Grouping Identical Objects for Region Proposal and Classification}

\author{Wim Abbeloos$^{1}$\thanks{Denotes joint first authorship}, Sergio Caccamo$^{2*}$, Esra Ataer-Cansizoglu$^{3}$, Yuichi Taguchi$^{3}$,\\ Chen Feng$^{3}$, and Teng-Yok Lee$^{3}$\\
\\
$^{1}$KU Leuven \quad $^{2}$KTH Royal Institute of Technology \quad $^{3}$Mitsubishi Electric Research Labs (MERL)} %
\maketitle

\begin{abstract}

\vspace{-0.25cm}
Often multiple instances of an object occur in the same scene, for example in a warehouse.  Unsupervised multi-instance object discovery algorithms are able to detect and identify such objects.  We use such an algorithm to provide object proposals to a convolutional neural network (CNN) based classifier.  This results in fewer regions to evaluate, compared to traditional region proposal algorithms.  Additionally, it enables using the joint probability of multiple instances of an object, resulting in improved classification accuracy.  The proposed technique can also split a single class into multiple sub-classes corresponding to the different object types, enabling hierarchical classification.

\end{abstract}

\vspace{-0.3cm}

\section{Introduction}
\label{sec:intro}

\begin{figure}[t]
\centering
\includegraphics[width=0.9\columnwidth]{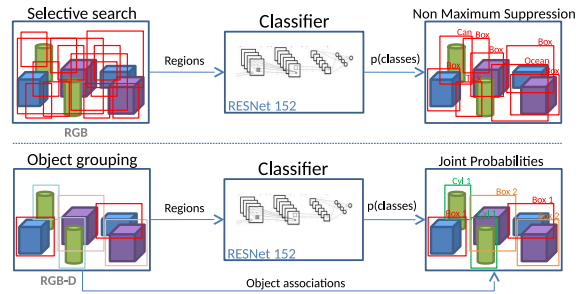}
\caption{A typical object detection pipeline utilizing a CNN (top), and the proposed pipeline using an alternative object discovery method for region proposal (bottom).}
\label{fig:system}
\end{figure}

Recent years have seen tremendous progress in object detection and classification based on convolutional neural networks (CNNs).  One of the standard pipelines is using an object proposal method and then classifying each region using a CNN~\cite{girshick2015fast}, as shown in Figure~\ref{fig:system} (top).  While general object proposal methods work well for general scenes, some scenes such as industrial and warehouse environments have more structure and the objects show less variations in appearance.

In this work, we exploit such structure by using a multi-instance object discovery algorithm~\cite{Abbeloos2017} that is able to discover, localize, and identify object instances that occur in a scene multiple times.  The algorithm uses an RGB-D image as the input and searches for patterns of local features that occur in multiple objects.  The output of the algorithm is clusters of the features, each corresponding to one instance of an object.  The clusters corresponding to multiple instances of the same object are associated with each other.
As shown in Figure~\ref{fig:system} (bottom), bounding boxes of the discovered objects are used as the input to a classification network. Our object proposal method also provides the identity of the object, allowing to accumulate the probabilities and further improve the classification accuracy.

\section{Experiments and Results}
\label{sec:exp}

\begin{figure*}[t]
\centering
\includegraphics[width=0.83\textwidth]{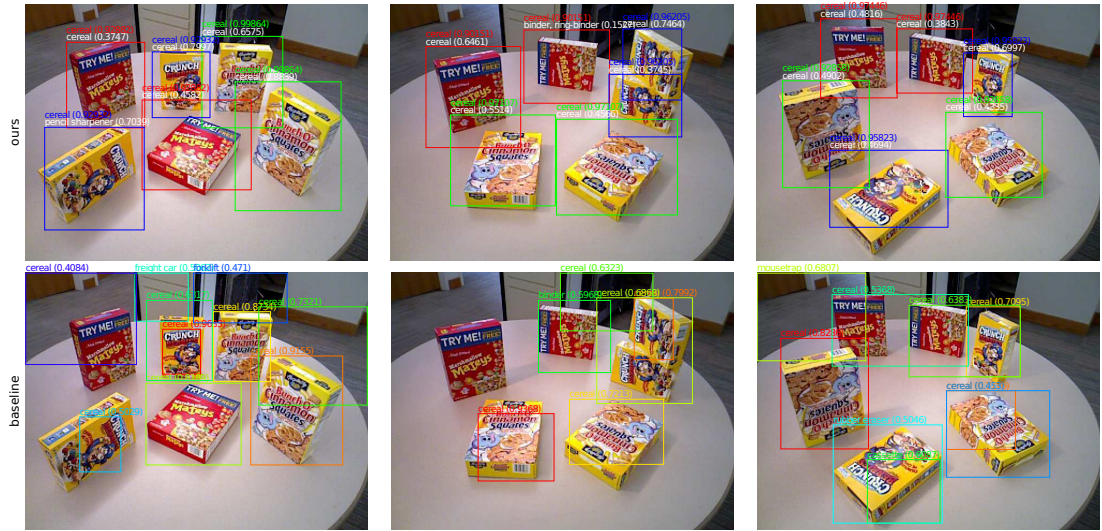}
\caption{Detection and classification results using our method (top) and the baseline method (bottom).  For our method the bounding box colors indicate object association.  We show both the individual (white) and joint (in color) probabilities.  For the baseline method we only show the regions with probability larger than $0.4$ for better visualization.}
\label{fig:result}
\end{figure*}

A dataset consisting of ten scenes with six cereal boxes on a table was captured using an ASUS XTion RGB-D camera.  There were three different object instances, two instances of each type (Figure~\ref{fig:result}).  We made sure the same faces of the objects of the same type were visible so that they could be discovered, identified, and localized properly.

For our method the bounding boxes were obtained by expanding each feature cluster by 60 percent to ensure they encapsulate the entire object.  The resulting regions were cropped from the images and resized to have 640 pixels on the longest side and a Gaussian filter was applied to reduce interpolation artifacts.  These images were then classified using ResNet-152~\cite{he2016deep}, pre-trained on ImageNet (1000 classes).  This resulted in a vector of 1000 class probabilities.

For the baseline method the bounding boxes returned by selective search~\cite{uijlings2013selective} were resized and classified using the same ResNet-152.  An extra non-maximum suppression step was used with a maximum overlap threshold of $0.5$.  A detection was considered correct if the intersection over union (IOU) was at least either $0.25$ or $0.50$.  The ground truth poses of the boxes were manually annotated.

\begin{figure}[t]
\centering
\includegraphics[width=0.31\textwidth]{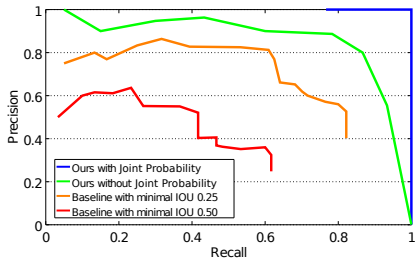}
\caption{The precision-recall curves for our method with (blue) and without (green) use of the joint probability.  The orange and red curves are the baseline method with minimal IOU set to $0.25$ and $0.50$ respectively.}
\label{fig:graph}
\end{figure}

The sum of the probabilities of the categories ``packet'' and ``comic'' was used as the probability of the object being a cereal box.  To evaluate the classification results, the precision and recall were calculated as
\begin{equation}\small
P =  \frac{ \sum_{i=1}^{C}{ tp_{i} }}{ \sum_{i=1}^{C}{ (tp_{i}+ fp_{i})}}, \quad
R = \frac{ \sum_{i=1}^{C}{ tp_{i} }}{ \sum_{i=1}^{C}{ (tp_{i}+ fn_{i})}},
\end{equation}
with $tp_{i}$, $fp_{i}$, and $fn_{i}$ the true positives, false positives, and false negatives, respectively.  These were summed over all  $C = 1000$ classes.  Note that a single incorrect classification might be counted as a false positive multiple times.

Figure~\ref{fig:graph} shows the precision-recall curves.  For our method we compared the result where we considered the objects as independent (without computing the joint probability), and that where we computed the joint probability of all instances of the same object.  The joint probability was computed by multiplying the class probability vectors of these instances and re-normalizing them, which increased the probabilities of correct classifications and improved the precision-recall performance.

Figure~\ref{fig:result} visually shows some of the scenes with the detected and classified objects.    As can be seen we have a much smaller number of detected regions, while the baseline method tends to generate many false positives.  The average number of proposed regions by our method was six, while that of the baseline method was $94$.  After non-maximum suppression the average number of remaining regions was $15$.

\section{Discussion}
\label{sec:discussion}
Our object detection framework proposes object regions based on discovering multiple instances of identical, feature rich objects.  While these are strong assumptions, several real-world robotic scenarios satisfy these conditions, which would benefit from our method.  Currently our evaluation was performed using a limited number of simplified scenes.  We plan to extend the evaluation using a larger number of more complex scenes.

\vspace{0.15cm}

\noindent {\bf Acknowledgment:}
W.A. and S.C. did this work during an internship at MERL.  W.A. thanks EAVISE, KU Leuven for the travel support.

{
\footnotesize
\bibliographystyle{ieee}
\bibliography{SectionsCR/HC}
}

\end{document}